\documentclass[letterpaper, 10 pt, journal, twoside]{IEEEtran}
\usepackage{amsmath,amsfonts}
\usepackage{algorithmic}
\usepackage{array}
\usepackage[caption=false,font=normalsize,labelfont=sf,textfont=sf]{subfig}
\usepackage{textcomp}
\usepackage{stfloats}
\usepackage{url}
\usepackage{verbatim}
\usepackage{graphicx}
\usepackage{diagbox}
\usepackage{makecell}
\usepackage{utfsym}
\usepackage[dvipsnames, svgnames, x11names]{xcolor}
\usepackage{hyperref}
\hypersetup{
	colorlinks=true,
	linkcolor=cyan,
	filecolor=blue,      
	urlcolor=blue,
	citecolor=green,
}
\def\BibTeX{{\rm B\kern-.05em{\sc i\kern-.025em b}\kern-.08em
    T\kern-.1667em\lower.7ex\hbox{E}\kern-.125emX}}
\usepackage{balance}

\begin{document}
\title{STAGE: STyle-controllable Action GEneration for personalized autonomous driving}
\author{Zihao Liu, Xing Liu$^{*}$, Yizhai Zhang, Panfeng Huang
\thanks{Manuscript received: April 30, 2025; Revised July 21, 2025; Accepted November, 8, 2025.}
\thanks{This paper was recommended for publication by Editor Tamim Asfour and Ki-Uk Kyung upon evaluation of the Associate Editor and Reviewers' comments. }
\thanks{This work was supported in part by the National Key R\&D Program of China under Grant 2022ZD0117903, and in part by Guangdong Major Project of Basic and Applied Basic Research under Grant 2023B0303000016, and the National Natural Science Foundation of China under Grant 92370123 and 62273280.}
\thanks{Zihao Liu, Xing Liu (Corresponding author), Yizhai Zhang, and Panfeng Huang are with the Research Center for Intelligent Robotics, School of Astronautics, Northwestern Polytechnical University, and National Key Laboratory of Aerospace Flight Dynamics, Northwestern Polytechnical University, Xi'an, China, 710072 e-mail: {\tt\small xingliu@nwpu.edu.cn}, {\tt\small pfhuang@nwpu.edu.cn}}
\thanks{Digital Object Identifier (DOI): see top of this page.}
}

\markboth{IEEE Robotics and Automation Letters. Preprint Version. Accepted November, 2025}
{Liu \MakeLowercase{\textit{et al.}}: STAGE} 

\maketitle

\begin{abstract}
Driving style refers to the behavioral preferences that drivers maintain during driving, shaped by their diverse experiences, habits, and needs, and is typically reflected in varying levels of aggressiveness. If humans choose to use autonomous driving systems, they would expect the driving style of the systems to closely resemble their own habit.
However, this is challenging for current industrial autonomous driving systems. To address this, we developed a style controllable action generation method, STAGE, for driving tasks. 
Its training process is based on imitation learning, incorporating both style value and latent value action modality encoding. Preference learning is then used to identify the user's driving style as a continuous, monotonic style value. And to reduce the cost of human involvement in the preference training process, we also developed a set of rules to compare driving style in data pairs. 
Then, during inference, the user inputs the style value to control the generated action patterns, dynamically meeting the user's expectations.
Using the STAGE method, we verified that the style-controlled action generation results in several typical road scenarios significantly align with human expectations. 
Furthermore, through comparisons between the STAGE method and various other approaches, we reveal the unique functionalities of STAGE, including its style
controllability, style continuity, driving style alignment capability and driving safety.
The code for this work is available at: 
\href{https://github.com/CarlDegio/STAGE}{github.com/CarlDegio/STAGE} 
\end{abstract}

\begin{IEEEkeywords}
Autonomous Vehicle Navigation, Imitation Learning, Human Factors and Human-in-the-Loop
\end{IEEEkeywords}

\section{INTRODUCTION}

\IEEEPARstart{A}{utonomous} driving technology has achieved remarkable progress, as exemplified by rule-based systems like Apollo \cite{apollo_em,apollo_qp} and end-to-end learning models such as UniAD \cite{uniad}, DriveGPT4 \cite{drivegpt4}. Simultaneously, Level 2 driving assistance systems are increasingly being implemented on roads, substantially reducing driver workload.
Despite these advancements, the current limitations of autonomous driving systems necessitate that drivers remain vigilant and prepared to assume control whenever the autonomous system is operational \cite{take_over}. 
This necessity results in a structure of human-machine collaboration, where both the human and the machine are involved in the driving process. Importantly, the machine’s driving behavior has a profound influence on the driver’s psychological and behavioral state.
For example, if autonomous vehicles exhibit behavior significantly different from that of human drivers, it can reduce user trust of the system, increasing the likelihood of drivers taking over control of the vehicle. \cite{drive_style_comfort}
In light of this, we propose that autonomous driving systems should integrate users' driving preferences and intentions to foster greater trust. Specifically, the driving behaviors generated by these systems should be designed to make users feel comfortable, closely resembling the experience of driving the vehicle themselves. 

\begin{figure}[t]
      \centering
      \includegraphics[scale=0.22]{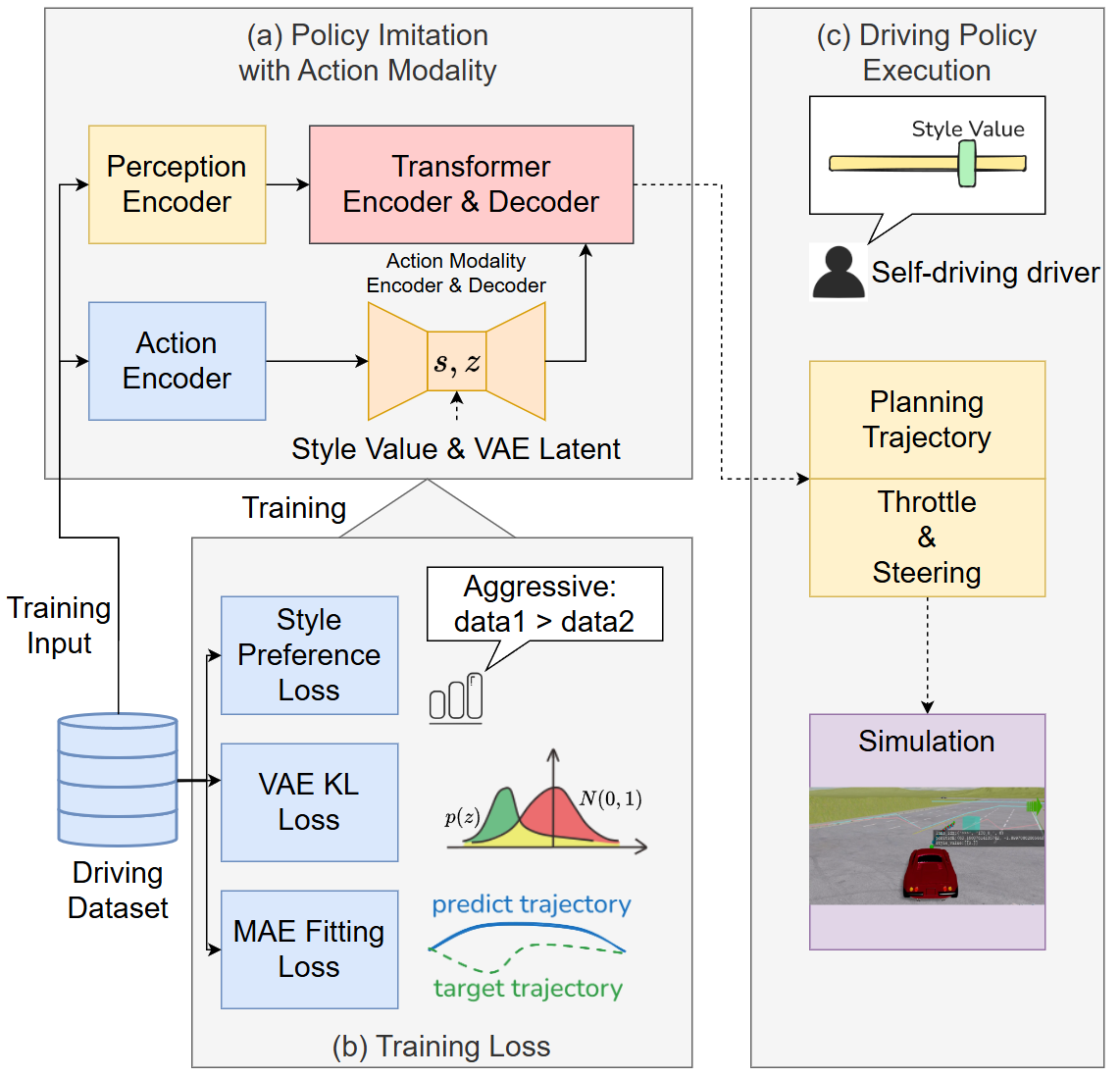}
      \caption{Overview of the STAGE method. The solid line arrows represent the data flow during the training process, and the dotted line arrows indicate the data flow during the deployment process. (a) shows the network architecture of the STAGE method, which employs a Transformer-based encoder-decoder structure for imitation learning. It extends the functionality of imitation learning with an action modality module consisting of a style encoder and a Variational Autoencoder (VAE). This module captures diverse action patterns in the driving dataset and generates driving actions that are consistent with the desired driving style. (b) describes the three types of losses used for training the model, where the style preference loss, Kullback-Leibler (KL) loss, Mean Absolute Error (MAE) loss, are employed to train the human preferences for driving aggressiveness, capturing action modalities and policy output respectively. 
      In (c), during testing, the autonomous driver can provide a style value by slide bar, which serves as a condition for generating both trajectories and vehicle control commands (steering, throttle), thereby meeting the driver's diverse needs.}
      \label{brief}
\end{figure}

User driving preferences can be characterized by driving styles, which denote variations in driving policies shaped by individual habits while adhering to safety standards. It is usually manifested as differences in driving aggressiveness such as habitual driving speed and lane change operations. 
Accurate analysis of a user's driving style is essential to generate behaviors that closely align with their individual driving patterns.

However, current methods for identifying driving styles have several limitations. Most existing approaches categorize driving behaviors into distinct classes \cite{style_classification_raw}, \cite{style_classification_LMKNN}, such as aggressive, cautious, and standard. While this classification is straightforward for supervised learning, it often fails to capture the nuances of individual users in practice. 
\textcolor{black}{Another category of approaches focuses on learning continuous latent representations using generative models, such as Variational Autoencoders (VAEs) \cite{VAE_compare1,VAE_compare2} or Generative Adversarial Networks (GANs) to represent driving styles \cite{style_implicit_gail}. These methods facilitate style modification by manipulating latent variables to control generated actions. However, the learned latent spaces in these works are often entangled and lack a monotonic relationship with specific semantic attributes (e.g., aggressiveness). While they allow for changing the style, they do not provide an intuitive "knob" where increasing the value linearly corresponds to an increase in a specific driving characteristic. Consequently, users cannot effectively adjust the latent vector to precisely achieve their desired level of aggressiveness due to this lack of semantic alignment.}

\textcolor{black}{Furthermore, recent advancements in Vision-Language-Action (VLA) driving models have attempted to control driving behaviors through natural language prompts. While language offers an intuitive interface, it suffers from inherent ambiguity and granularity issues in the context of vehicle control. For instance, the command "drive a bit faster" is subjective; its execution varies significantly depending on the underlying model and context, lacking a precise definition. In contrast, driving style—specifically aggressiveness—is better modeled as a continuous spectrum. }

To address these issues, we propose a preference-based style learning method. \textcolor{black}{Unlike standard VAE approaches that learn unstructured latent spaces, our method imposes a preference constraint on driving style latent dimension.} It extracts driving styles from the action modalities of the policy, thereby providing an accurate, continuous, and trend-consistent numerical representation of driving styles. The overall architecture of the method is shown in Figure \ref{brief}.
First, in the network architecture, an action modality encoder was designed, taking both state and action as input to extract the multi-valued distribution caused by differences in driving styles from the driving dataset. This distribution reflects the existence of multiple distinct actions under the same state, collectively referred to as action modalities.
And during training, to minimize human annotation involvement, our method leverages aggressive scoring to compare randomly selected action pairs, capturing human preference information across diverse scenarios and driving trajectories. The trained model is then capable of evaluating the \textbf{style value} of a given planning trajectory and vehicle control in a specific state, effectively quantifying the level of driving aggressiveness. This approach ensures a monotonic relationship between style value and driving aggressiveness, where higher values correspond to more aggressive driving behaviors, thereby aligning the style value with human driving preferences. 
Furthermore, the style value can be utilized as a conditional input in imitation learning to generate corresponding driving actions. To achieve this, we propose a STyle-controllable Action GEneration (STAGE) network, which combines an action modality module with a Transformer backbone. This network leverages the style value and multimodal environmental states as conditions to imitate the driving policies present in the dataset. During deployment, it allows drivers to adjust the behavior of the autonomous driving system by inputting a style value until it meets their preferences, thereby enhancing the comfort and personalization of autonomous driving.

To summarize our contributions in this work:
\begin{itemize}
\item We designed an action modality module using an encoder-decoder architecture to process the input state and action, extracting a style value that represents driving aggressiveness and a VAE latent that captures the remaining policy patterns. 
\item We propose a preference learning method to learn driving style values, delivering continuous and consistent assessments of driving behaviors across various scenarios. This approach overcomes the limitations of previous methods, which could only provide discrete driving styles. Additionally, we designed rules that use aggressiveness scoring to automate preference comparisons, reducing the manual effort required for preference learning.
\item We also designed a style-conditioned action generation method that incorporates both policy imitation and a user intervention mechanism. Users can directly input their desired style value during model execution, enabling the policy to generate outputs consistent with the specified style. 
\end{itemize}

\section{RELATED WORK}
\subsection{Imitation Learning}
In recent years, with the advancement of neural network technology, imitation learning that fits policies from demonstration data has seen new development opportunities. Behavioral Cloning (BC) \cite{behavior_cloning1,behavior_cloning2} is a fundamental imitation learning method that directly fits the policy $\pi(a|s)$ and is readily implemented using neural network architectures. Thus, it has found widespread application in various domains, such as autonomous driving \cite{behavior_cloning_drive} and robotic skills transfer \cite{behavior_cloning_robot}. However, it suffers from issues such as error accumulation, multi-modality confusion, and sometimes difficulty in fitting the policy. 
To address these issues, another class of imitation learning methods based on state-value assessment, known as Inverse Reinforcement Learning (IRL) \cite{IRL}, has been proposed. Subsequent developments include Generative Adversarial Imitation Learning (GAIL) \cite{GAIL}, which uses generative adversarial networks for the action generator and reward discriminator, and the decision transformer (DT) \cite{DTrobot}, which employs sequence modeling of task trajectories. The BC and IRL constitute the two major branches in the field of imitation learning.

A new development direction in imitation learning is the use of scene vision and embodied vision to provide observational information, which can avoid the need for unique state space designs for various tasks. However, this also imposes new requirements on learning algorithms. For example, implicit behavior cloning (IBC) \cite{IBC} uses an energy-based model to perform implicit regression to predict future actions, achieving better results in high-dimensional action spaces and with visual image inputs.
Recently published algorithms, such as Action Chunking Transformer (ACT) \cite{aloha1} and Diffusion Policy \cite{diffusion_policy}, represent the state of the art in the field of visual imitation learning. The former leverages the multimodal compatibility of transformers combined with Conditional-VAE (CVAE) design to achieve skill learning from small datasets, while the latter, based on the diffusion model, proposes a denoising approach to fit action sequences.
In the field of autonomous driving, similar ideas have been applied to end-to-end driving, where driving policies are generated without the need for a HD map. The UniAD \cite{uniad} method outlines the paradigm for end-to-end autonomous driving, introducing a pipeline that requires only visual information. It is planning-oriented and employs collaborative learning of perception, prediction, and planning, showing excellent results across various outputs. 
Therefore, our approach uses the characteristics of end-to-end driving and imitation learning, utilizing a transformer architecture to learn trajectory planning.

\subsection{Driving Style}
Overall, current methods for recognizing driving style can be divided into two categories: statistical learning-based and deep learning-based. Statistical learning methods typically use techniques like Principal Component Analysis (PCA) and Support Vector Machine (SVM) to transform driving style recognition into a classification problem based on driving states. For example, using PCA, driving styles can be classified as aggressive, moderate, and conservative, guiding the Intelligent Driver Model (IDM) strategy to generate three types of policies \cite{car_follow_style}. Other methods model the driving process using the Mixture of the Hidden Markov Model (MHMM) \cite{MHMM_style}, ensuring smooth transitions in style recognition. SVM, as an efficient classification method, can be combined with human-designed rules for low-cost style recognition \cite{SVM_style_hardware}.

In deep learning-based driving style recognition, the methods can be categorized as explicit or implicit. Explicit methods treat driving style as a classification problem, such as classifying the driving styles of other vehicles and using them as input to the vehicle policy to help in better decision making \cite{DETR_moe_style_traj}. Implicit style recognition is more beneficial for generating the trajectory of the ego vehicle. For example, statistical features of driving behavior over time can represent driving style, forming a driving operational picture (DOP). A neural network then encodes the DOP as part of the policy input to generate human-like lane-changing strategies \cite{DOP_style}. 
Furthermore, the GAIL method can learn implicit representations of driving styles \cite{style_implicit_gail} and generate new trajectories.

However, we observe that classification-based methods cannot precisely align with driver intent, while implicit style representations make it difficult for drivers to directly adjust the generated trajectory. Our approach addresses these issues by incorporating a continuous style value that corresponds monotonically with the driving aggressiveness.

\section{METHOD}
The STAGE method consists of two parts: preference-based style learning and style conditioned action generation. In the style learning part, we first designed a preference-based comparison method for driving data pairs, enabling the model to learn continuous and consistent style values. Subsequently, we constructed our imitation learning network, which takes environmental states and human actions as inputs to extract environmental features and action modalities, while separating the style value and training it using preference. The action modality and environmental state are then fed into the transformer decoder to generate customized actions. The advantage of the STAGE method lies in its ability to allow users to specify style values in real-time during testing, enabling the generation of the model consistent with the specified driving style.

\subsection{Style Learning from Pairwise Preference}

Previous methods for style recognition relied on manually labeled supervised classification learning. We believe that this approach hinders alignment between human expectations and driving behaviors, as well as the generation of diverse style trajectories. In contrast, continuous driving style values are more suitable for representing driving styles. To address this, STAGE uses preference learning to output continuous style values. 
Additionally, other methods typically equate driving style with driving aggressiveness, whereas the STAGE method can derive style values representing richer driving characteristics through preference learning, enabling influence on the behavior of autonomous driving systems across multiple dimensions. However, in the current paper, we align with existing driving style literature by primarily focusing on a one-dimensional style value aligned with driving aggressiveness, where a higher style value indicates more aggressive driving behavior.

Specifically, preference learning formalizes the style learning problem as follows:

\begin{equation}
    P((x^+,a^+) \succ (x^-,a^-))=\frac{e^{V(x^+,a^+)}}{e^{V(x^+,a^+)}+e^{V(x^-,a^-)}}
\end{equation}

Here, $P((x^+,a^+) \succ (x^-,a^-))$ denotes the probability that taking action $a^+$ (the planning trajectory and control signal) in state $x^+$ is more aggressive than taking action $a^-$ in state $x^-$, and $V(x, a)$ is used to evaluate the style value of taking action $a$ in state $x$, representing the level of aggressiveness. The training of the style value is conducted by obtaining the preference ranking between different $(x, a)$ pairs.
It is noteworthy that the style value $V(x,a)$ here can be directly extended to a multidimensional vector, thereby defining additional driving style characteristics on the $(x,a)$ pair. However, this requires more data and annotation support.



Although human involvement can achieve this, the need for human annotation during the training process results in excessive labor costs. To address this, we opt for preference comparison rules as a low-cost alternative that offers similar performance.
Specifically, samples from dataset were preprocessed into structured information. This includes the control signal of the ego vehicle (the self vehicle in the autonomous driving system), its lane deviation, and speed, as well as its relationship with other vehicles (such as changes in following distance or overtaking other vehicles).
Then, this information is input into an aggressiveness scoring function for evaluation, allowing comparison of which action in a pair is more aggressive. 
We developed a generalized aggressiveness scoring rule, where the speed and throttle amplitude of the ego vehicle are proportional to the aggressiveness score. When the nearest other vehicle is within 20 meters, two additional scoring terms are considered: one inversely proportional to the distance to the nearest vehicle, indicating that closer proximity to other vehicles reflects more aggressive driving, and another proportional to the deviation from the lane line, signifying that overtaking other vehicles represents aggressive driving. Notably, our style value model, trained through preference learning, aligns with these rules. However, potential users can modify the alignment of style values by adjusting the rule or utilizing human annotated datasets.
In Figure \ref{train} describes this process.


Thus, the loss of preference learning can be expressed as the negative logarithmic likelihood of the probabilities, which will be part of the overall loss:

\begin{equation}
    \mathcal{L}_p=
    -\mathbb{E}_{(x^+,a^+),(x^-,a^-)\sim
    \mathcal{D}}  
    \log
    \left[
    \frac{e^{V(x^+,y^+)}}{e^{V(x^+,y^+)}+e^{V(x^-,y^-)}}
    \right]
\end{equation}
In this context, \(\mathbb{E}_{(x^+,a^+),(x^-,a^-)\sim\mathcal{D}}\) represents the expectation over pairs of data \((x^+, a^+)\) and \((x^-, a^-)\) sampled from the dataset \(\mathcal{D}\). 

\subsection{Stylized Action Generation Framework}

\begin{figure}[thpb]
      \centering
      \includegraphics[scale=0.18]{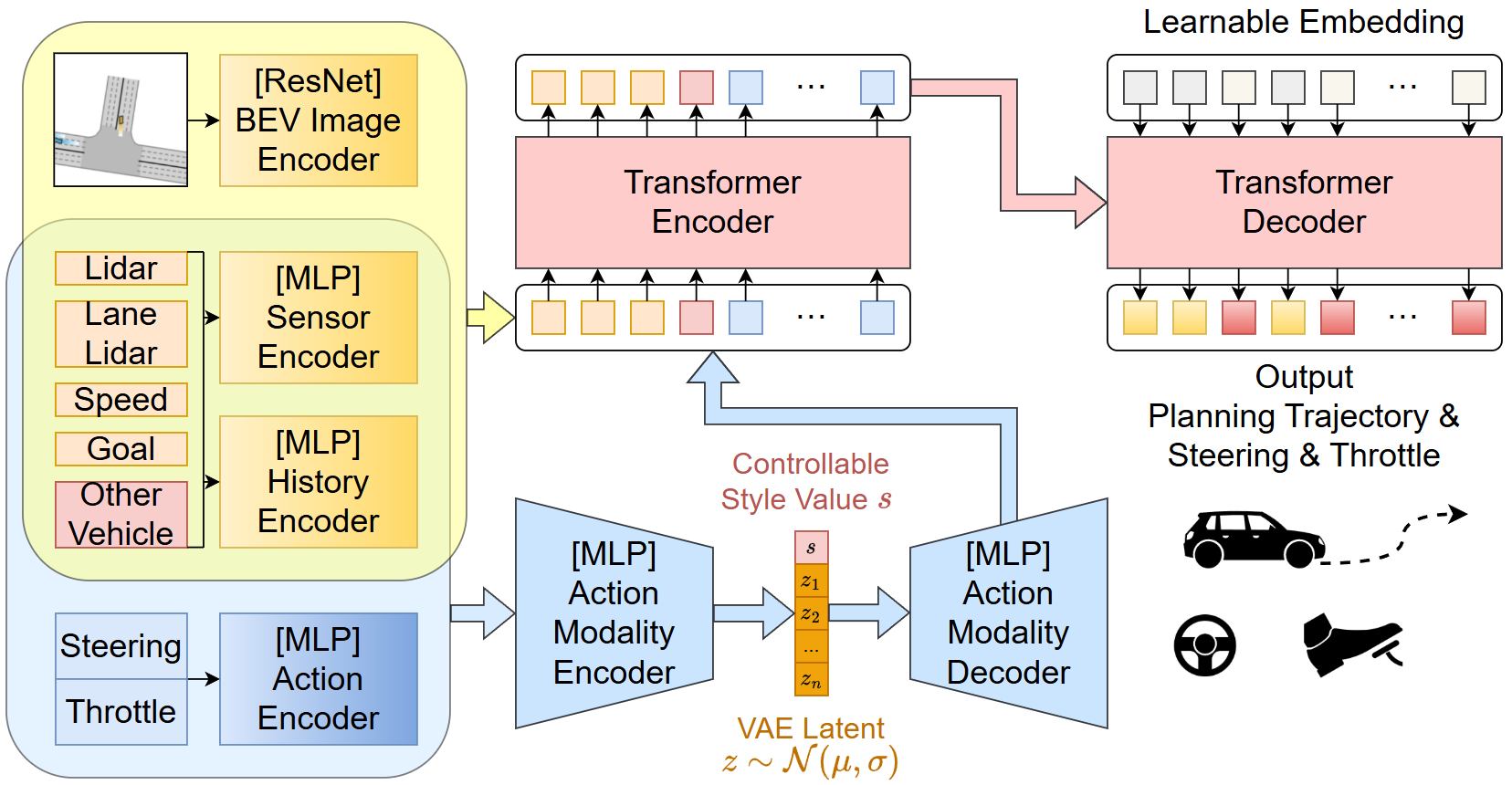}
      \caption{A network architecture diagram of the STAGE method. The action modality learning module utilizes a network architecture similar to compression encoding, while simultaneously training both preferences and VAE latent variables. The actions imitation adopts a Transformer architecture with multimodal inputs (based on DETR architecture) to extract features and output customized driving actions.}
      \label{network}
\end{figure}

Inspired by recent advances in efficient imitation learning based on Transformers, we propose an effective imitation learning model, which combines action modality learning, style preference supervision, and action supervision, built upon transformer-based supervised learning. 
The model leverages the multimodal information fusion capabilities of the Transformer encoder, allowing the simultaneous input of map images, the history states of the ego vehicle and other vehicles, as well as sensor data from the ego vehicle for feature extraction. 
At the same time, the action modality encoder will encode the current state and the actions taken by humans in the dataset at that state, representing the diverse patterns reflected in the human policy (such as different behaviors in the same state). And, this action modality information is also fed as input to the transformer encoder. 
Finally, after processing by the encoder, the transformer decoder generates a future trajectory that aligns with the desired driving style and outputs the corresponding steering and throttle control signals for vehicle execution.
The overall network architecture diagram of the STAGE method is shown in Figure \ref{network}.

In the transformer encoder, to accommodate multimodal input, image information is encoded using ResNet to form several token embeddings, while vector information (such as radar scan points and the historical trajectories of the ego vehicle and other vehicles) is directly mapped into tokens via an 
Multilayer Perceptron (MLP) network. 
Subsequently, the driving style and randomness of driving behavior extracted from the action modality are also embedded as a token, which together with the previous information constitutes the input token sequence of the transformer encoder.
Finally, the transformer encoder extracts features of the environmental state through the self-attention mechanism.

Next, we focus on the action modalities within the dataset. Action modalities refer to the multi-valued distribution of state-action pairs in the collected dataset, a phenomenon frequently observed in human driving behavior. This arises because humans exhibit diverse driving patterns influenced by habits or situational demands. We observed a close correlation between action modalities and driving styles: action modalities encompass driving styles but also include other patterns unrelated to driving styles. For instance, when driving on an open roundabout, the driving speed reflects the level of aggressiveness and is thus related to driving style, whereas the steering angle adjustments are merely to maintain a circular trajectory within the roundabout, independent of driving style. This separability indicates that the previously mentioned style preference learning is insufficient to fully capture driving behavior. Therefore, we need to separately extract two components from the action modalities: driving style and general behavioral patterns. To this end, we designed a two-part approach in the compression encoding of the MLP-based action module. The first part aligns with human understanding of driving style, using style values derived from preference learning to represent the driving aggressiveness. The second part aims to capture the remaining behavioral patterns, employing latent variables from a variational autoencoder (VAE) to represent the diversity of driving behaviors. The combined effect of these two components enables the model to fully leverage the diversity within the dataset while enhancing its ability to intuitively align with human driving styles.

This design introduces a second loss function based on VAE, further enhancing the model's ability to replicate human-like driving behaviors:

\begin{equation}
    \mathcal{L}_{vae}=
    \mathbb{E}_{(x,a)\sim
    \mathcal{D}}  
    \mathbb{KL}(q_\theta(z|x,a)||\mathcal{N}(0,I))
\end{equation}
Here, \(\mathbb{KL}\) represents the Kullback-Leibler (KL) divergence, which measures the distance between the posterior distribution \(q_\theta(z|x,a)\) and the prior distribution \(\mathcal{N}(0, I)\). And \(q_\theta\) refers to the network that encodes the latent variables.

This structure for action modality offers an additional benefit: it enables controllable generation. During execution, the action modality encoder is not required. For VAE latent variables, zero will be fed during generation to retrieve the most well-trained actions, enhancing the stability of the policy output. As for the style value, it will be manually provided by the user; larger inputs will lead to more aggressive driving behaviors, aligning the actions with the preferences of users.
In summary, this structure can be viewed as one in which the state of the environment determines the base policy, while the user can intervene in the policy by changing the style value, tailoring the action modalities to their needs and thus improving the autonomous driving experience.

\begin{figure}[thpb]
      \centering
      \includegraphics[scale=0.17]{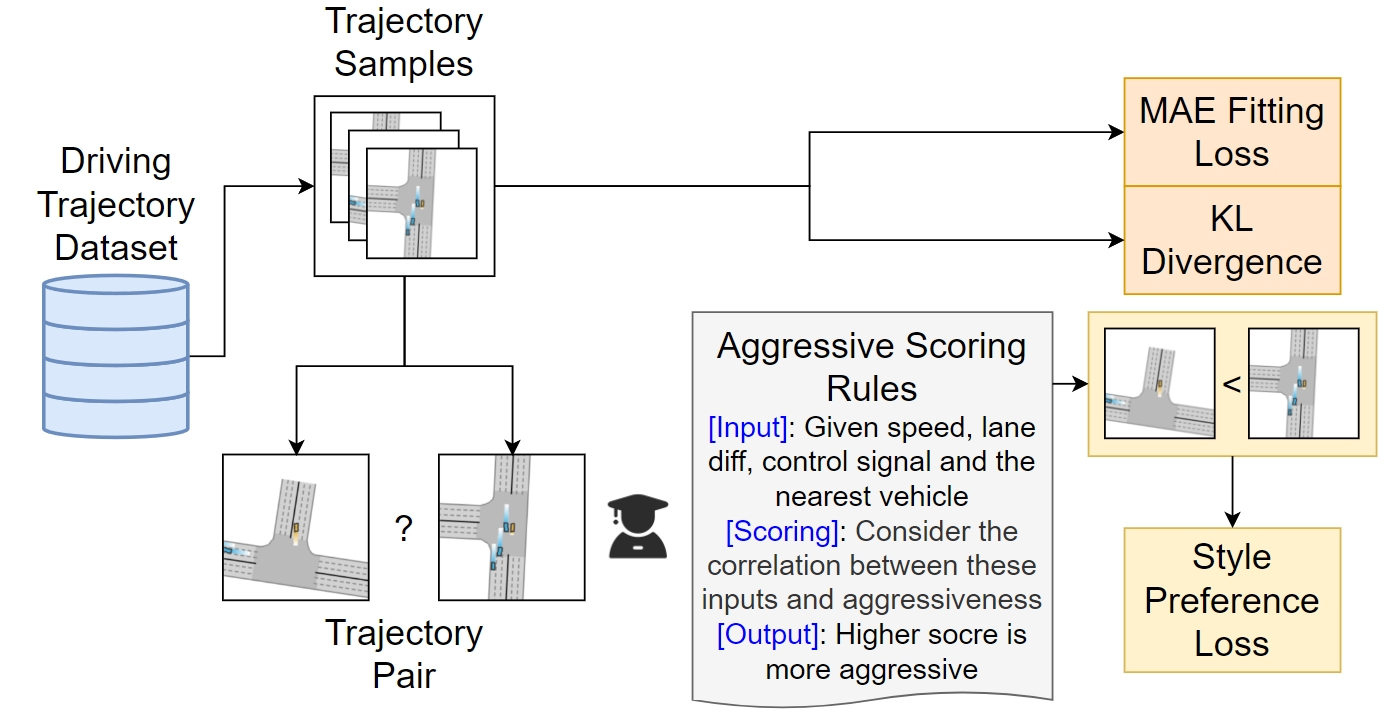}
      \caption{Composition of the loss function. The overall loss consists of three components: the MAE fitting loss for learning the output trajectory and control signals, the KL divergence for capturing general action modalities, and the style preference loss based on aggressive scoring rules which compares the driving aggressiveness between pairs of actions.}
      \label{train}
\end{figure}

In using the Transformer decoder to output the policy, we choose to output the planning trajectory, steering, throttle, and brake commands for the ego vehicle as actions. 
To achieve this, for each sampling moment in the dataset, both the control signals at that moment and the future position of the ego vehicle are used as supervision for training. After training, the policy will predict the ego vehicle's current control signals and future position. The control signals are then executed by a simulation, and the future trajectory is visualized. This leads to a loss function for action fitting:

\begin{equation}
    \mathcal{L}_{mae}=
    \mathbb{E}_{(x,a)\sim
    \mathcal{D}}  
    ||\hat{a}-a||_1
\end{equation}
Here, \(a\) is extracted from the dataset, containing both the trajectory and the control signals, while \(\hat{a}\) represents the predicted actions from the model.

Therefore, the update of the neural network can be performed in an end-to-end manner, and the overall loss function can be expressed as follows:
\begin{equation}
    \mathcal{L}=\mathcal{L}_{mae}+\lambda_1 \mathcal{L}_p + \lambda_2\mathcal{L}_{vae}
\end{equation}
Here, $\lambda_1, \lambda_2$ is the weight coefficient for the preference learning and VAE latent. A simplified diagram of the training process is shown in Figure \ref{train}.

\section{Experiments and Evaluation}

\begin{figure}[thpb]
      \centering
      \includegraphics[scale=0.11]{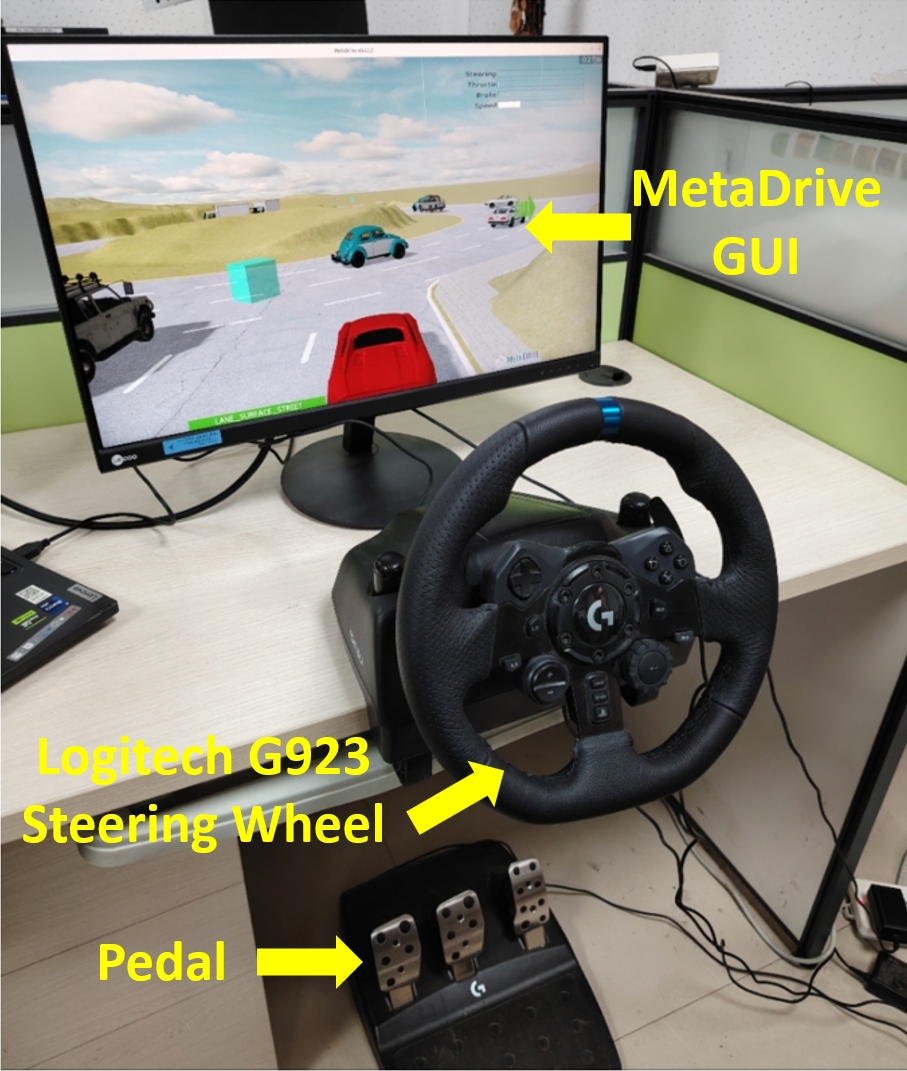}
      \caption{Our simulation engine utilizes the fast and convenient MetaDrive, with data collection conducted through an Logitech G923 steering wheel and pedal. Additionally, user style value inputs during testing are provided via a slider implemented by PySimpleGUI library.}
      \label{logitech_device}
\end{figure}


Considering the potential risks of implementing highly aggressive driving in physical vehicles, we collected diverse driving data within a driving simulator. We utilized MetaDrive \cite{metadrive}, a lightweight driving simulator capable of generating diverse map roads and traffic flows, thus ensuring the dataset covers a wide range of scenarios. MetaDrive also employs the Intelligent Driver Model (IDM) for other vehicles, making that all trajectories in the dataset are generated in a closed-loop manner, which is valuable for reflecting real-world driving conditions.

To record various human driving behaviors for preference comparisons and closely mimic real-world driving scenarios, we used the Logitech G923 steering wheel and pedal set as input for the ego vehicle. The rotation angles of the steering wheel and the throttle and brake of the pedal were assigned to the simulation system. This semiphysical setup allows human participants to behave as realistically as possible during data recording, thereby minimizing the disparity between the simulation system and real-world driving. This system is demonstrated in Figure \ref{logitech_device}.

We instructed drivers collecting data to deliberately exhibit diverse driving styles in the simulation to cover potential behaviors typical of everyday human driving. For example, during an unprotected left turn, drivers might either proceed without slowing down despite oncoming traffic or wait at the intersection until other vehicles have passed. Similarly, on a straight road, drivers might choose to follow a slow-moving vehicle or overtake it. This diversity in behavior increases the likelihood of generating effective comparisons in preference learning, thus aiding in the learning of style values. Moreover, to facilitate testing and comparison of experimental results, the style values of the trained model were scaled to the range \([-1, 1]\). A style value of 0 represents the regular driving behaviors, while negative values indicate conservative behavior and positive values represent aggressive behavior. Thus, larger style values correspond to more aggressive behavior specifications.

To validate the performance of our algorithm and compare it with other methods, we conducted three sets of experiments: (A) \textbf{Style Value Behavior Evaluation} where trajectory planning results of the STAGE method in representative road scenarios are analyzed. This experiment verifies that the model's behavior is controlled by the style value and qualitative aligned with human intent; 
(B) \textbf{Style Value Alignment Evaluation}, which compares the functionality of STAGE with various advanced methods in terms of style values and contrasts the style alignment results of STAGE with discrete style trajectory generation strategies. The results demonstrate that our method achieves the best driving style alignment capability; 
(C) \textbf{Safety Evaluation}, which analyzes the enhancement of driving safety through action modality learning, based on the average safe driving distance and the aggressiveness of throttle control commands.

\begin{figure}[thpb]
      \centering
      \includegraphics[scale=0.23]{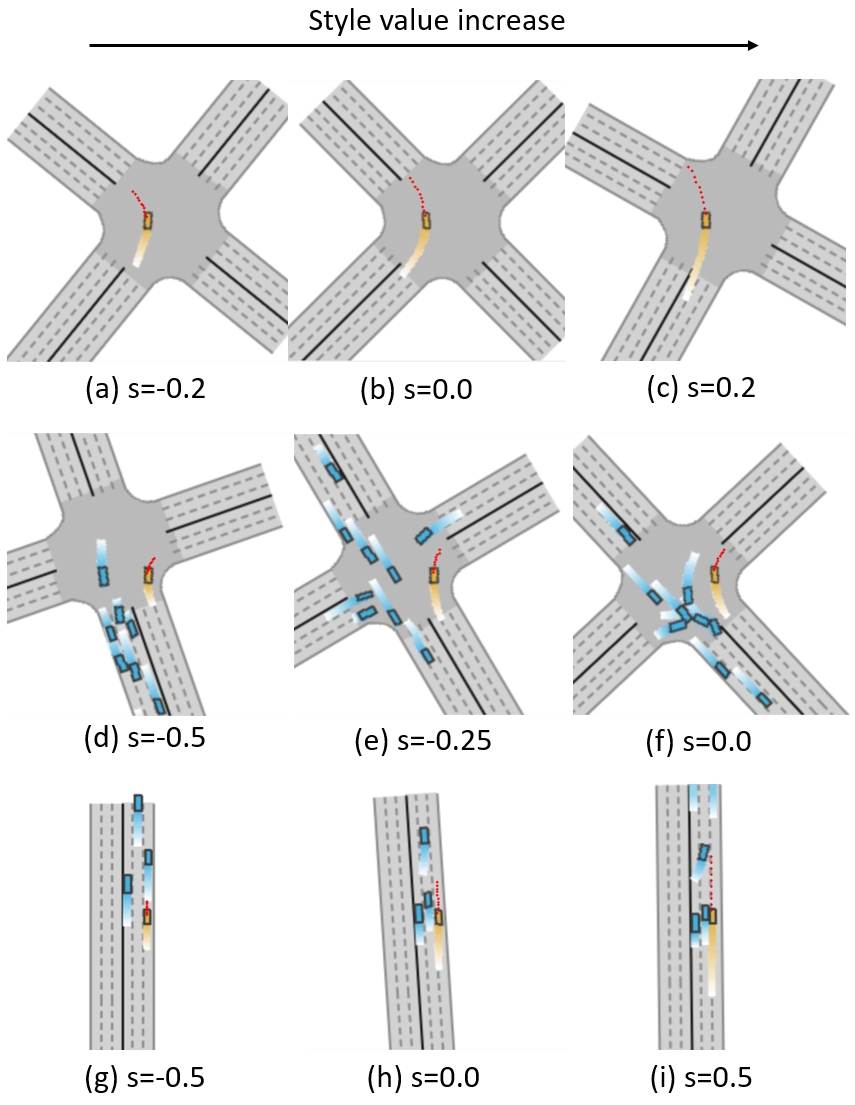}
      \caption{The results of the STAGE model in three typical scenarios are shown. The "s" following the subheading represents the style value used in this case. In each figure, the yellow rectangle represents the position of the ego vehicle, blue boxes represent the positions of other vehicles, and the respective trails indicate the historical movements of the vehicles. The red dotted line represents the planned trajectory of the ego vehicle over the next 1 second, and a longer dotted line indicates that the ego vehicle will have a higher speed in the future. From left to right, the style value setting increases sequentially, and from top to bottom, the scenarios are as follows: left turn at an intersection, right turn at an intersection, and overtaking on a straight road.}
      \label{scene}
\end{figure}

\subsection{Style Value Behavior Evaluation}
We demonstrated the STAGE algorithm across several representative scenarios generated by simulator, including straight driving, left and right turns at intersections, as shown in Figure \ref{scene}. Within these scenarios, we showcased how users can generate different behavior planning for the ego vehicle by adjusting the style value. 
Furthermore, since the traffic flow in these scenarios is generated randomly, the autonomous driving behaviors observed here further indicate that our method possesses a certain level of generalizability.

A comparison of the results in (a), (b), and (c) reveals that in the unprotected left-turn scenario, the style value directly influences the length of the planned trajectory. Larger style values result in trajectories with higher future speeds, aligning with human understanding of driving aggressiveness. 
Additionally, in Figures (d), (e), and (f), it can be observed that different style values sometimes influence the lane into which the vehicle turns during a right turn. We believe this is because the model considers the motion capability constraints allowed by the vehicle's speed when making decisions, thereby reflecting the potential relationship between the style value and driving aggressiveness. And, due to the presence of numerous nearby vehicles, the generated trajectories are relatively conservative to avoid collisions with surrounding vehicles.
Finally, in Figures (g), (h), and (i), the policy with low style values exhibits characteristics such as maintaining the same speed as other vehicles or waiting behind for other vehicles to pass. In contrast, the policy with high style values generates trajectories that overtake other vehicles. This indicates that higher style values result in more aggressive driving behaviors when interacting with other vehicles.

In summary, these validations demonstrate that the STAGE has the capability to generate style trajectories that align with human user understand of aggressiveness. This makes autonomous driving behaviors more consistent with human intentions, thereby reducing distrust in autonomous driving systems.

\subsection{Style Value Alignment Evaluation}

We validate the uniqueness of the STAGE method by comparing its functionality with various advanced approaches. Specifically, we include several methods from the field of driving style policy generation, such as DETR-based Behavioral Cloning (BC) \cite{DETR_BC}, Generative Adversarial Imitation Learning (GAIL) \cite{style_implicit_gail}, Conditional Variational Autoencoder (CVAE) \cite{cvae_driving} , CVAE with Discrete Style (CVAE+Discrete Style) \cite{cvae_classifier_style, DETR_moe_style_traj}, and a variant of the STAGE method constructed by removing the CVAE component: Imitation Learning with Preference-Based Style (BC+Preference Style). Their functional comparison is presented in Table \ref{sota_comparison}. The table evaluates these methods based on the controllability and continuity of driving styles. We find that existing methods fail to satisfy both criteria simultaneously, thereby hindering progress in style driving trajectory generation. In contrast, the proposed STAGE method and its variant achieve both functionalities concurrently, revealing the significant potential of applying preference learning to driving style learning. 

\begin{table}[h]
\caption{Comparison of STAGE with other representative methods}
\label{sota_comparison}
\begin{center}
\begin{tabular}{|c|c|c|c|}
\hline
\diagbox{Methods}{Metrics} & \makecell{Style \\ Controllability} & \makecell {Style \\ Continuity} & \makecell{Average \\Completion \\ Rate/\%} \\
\hline
BC(Based on DETR) & \textcolor{red}{\usym{2718}} & \textcolor{red}{\usym{2718}} & 51.0 ($\pm$22.8)\\
\hline
GAIL & \textcolor{red}{\usym{2718}} & \textcolor{Green}{\usym{2714}} & 42.6 ($\pm28.8$) \\ 
\hline
CVAE & \textcolor{red}{\usym{2718}} & \textcolor{Green}{\usym{2714}} & 83.4 ($\pm15.4$)\\ 
\hline
CVAE+Discrete Style & \textcolor{Green}{\usym{2714}} & \textcolor{red}{\usym{2718}} & 78.6 ($\pm$ 28.6)  \\
\hline
BC+Preference Style & \textcolor{Green}{\usym{2714}} & \textcolor{Green}{\usym{2714}} & 62.1($\pm$25.1) \\
\hline
\textbf{STAGE(ours)} & \textcolor{Green}{\usym{2714}} & \textcolor{Green}{\usym{2714}} & \textbf{92.1 ($\pm$ 20.1)} \\
\hline
Expert Driver & - & - & 100.0 \\
\hline
\end{tabular}
\end{center}
\end{table}

\begin{figure}[thpb]
      \centering
      \includegraphics[scale=0.2]{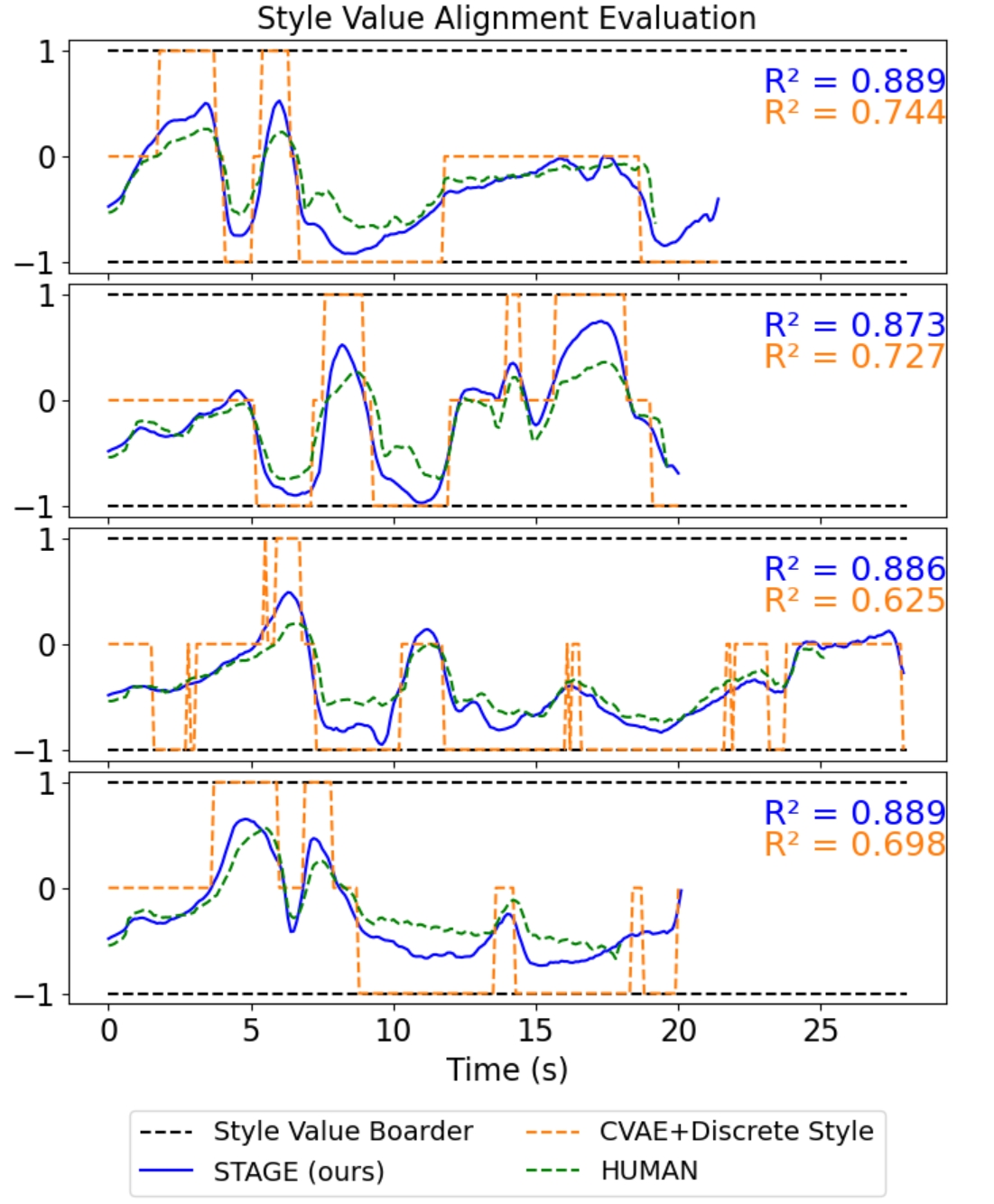}
      \caption{The style value alignment comparison results are presented. The style value sequence of the human policy (green dashed line) was compared to the style value sequence of the STAGE policy (blue solid line) and CVAE+Discrete Style policy (orange dashed line). Spearman correlation coefficient $R^2$ is used to measure the extent to which the two methods approximate human driving styles.}
      \label{style_value_eval}
\end{figure}

To further compare the effectiveness of style alignment, we contrast STAGE with a representative discrete style trajectory generation strategy. Initially, human driving data is collected, and the style value sequence is calculated based on STAGE's style value module, serving as the style benchmark. Subsequently, this sequence is input into both the STAGE model and the discrete style CVAE model to generate new driving trajectories. The style values of the generated trajectories are then evaluated using their respective style value modules as the outcome of driving style imitation. In this process, the style values of the STAGE method are normalized to the range [-1, 1], while the discrete three-category style values (conservative, moderate, and aggressive) are mapped to {-1, 0, 1}. Finally, the Spearman correlation coefficient ($R^2$) is used as a measure of similarity to the style value benchmark, where a value closer to 1 indicates better style imitation performance. Figure \ref{style_value_eval} presents the experimental results of style value alignment, showing that the $R^2$ of the STAGE is closer to 1, demonstrating a superior reproduction of human intent style values. 
Additionally, discrete styles often exhibit short-term abrupt changes, leading to instability in the driving process.
This suggests that continuous style representation facilitates the alignment of autonomous driving style with human driving styles.



\subsection{Safety Evaluation}

We will use the safety evaluation of the different policy to illustrate the advantages of our proposed action modality architecture. We defined the Average Completion Rate metric, which represents the mean and standard deviation of the percentage of mileage safely completed by the policy across 20 randomly generated driving scenarios in simulation. Table \ref{sota_comparison} lists the performance results of various methods, with BC and GAIL methods scoring the lowest due to their lack of explicit learning of action modalities. Next is the preference-based BC, which captures some action modalities based on aggressiveness scoring rules but misses style-independent action modalities, resulting in suboptimal driving safety. Methods incorporating the CVAE architecture (CVAE, CVAE+Discrete Style, and STAGE) achieve the best results, indicating that the CVAE architecture excels at capturing the action distribution in driving datasets. Furthermore, for methods that allow human driver intervention in the driving process, STAGE and CVAE+Discrete Style achieve similarly superior performance. This highlights the role of human drivers in autonomous driving systems, enhancing safety by controlling the driving style dynamically.


On the other hand, statistics of control signals can demonstrate that style learning within action modalities enhances driving safety by improving driving comfort. Figure 7 presents the statistics of throttle and brake usage during driving. It is evident that GAIL, BC, and CVAE, which lack driving style learning, exhibit a higher proportion of heavy throttle behavior. Additionally, the BC and CVAE methods show the most frequent braking actions, indicating poor learning of action modalities and a tendency to switch between various action modalities, resulting in more frequent acceleration and deceleration behaviors. In contrast, the STAGE method with continuous style values and BC+Preference Style both exhibit fewer braking actions and less extreme acceleration. They typically employ light throttle for driving, thereby enhancing the driver's experience and improving driving safety.


\begin{figure}[thpb]
      \centering
      \includegraphics[scale=0.24]{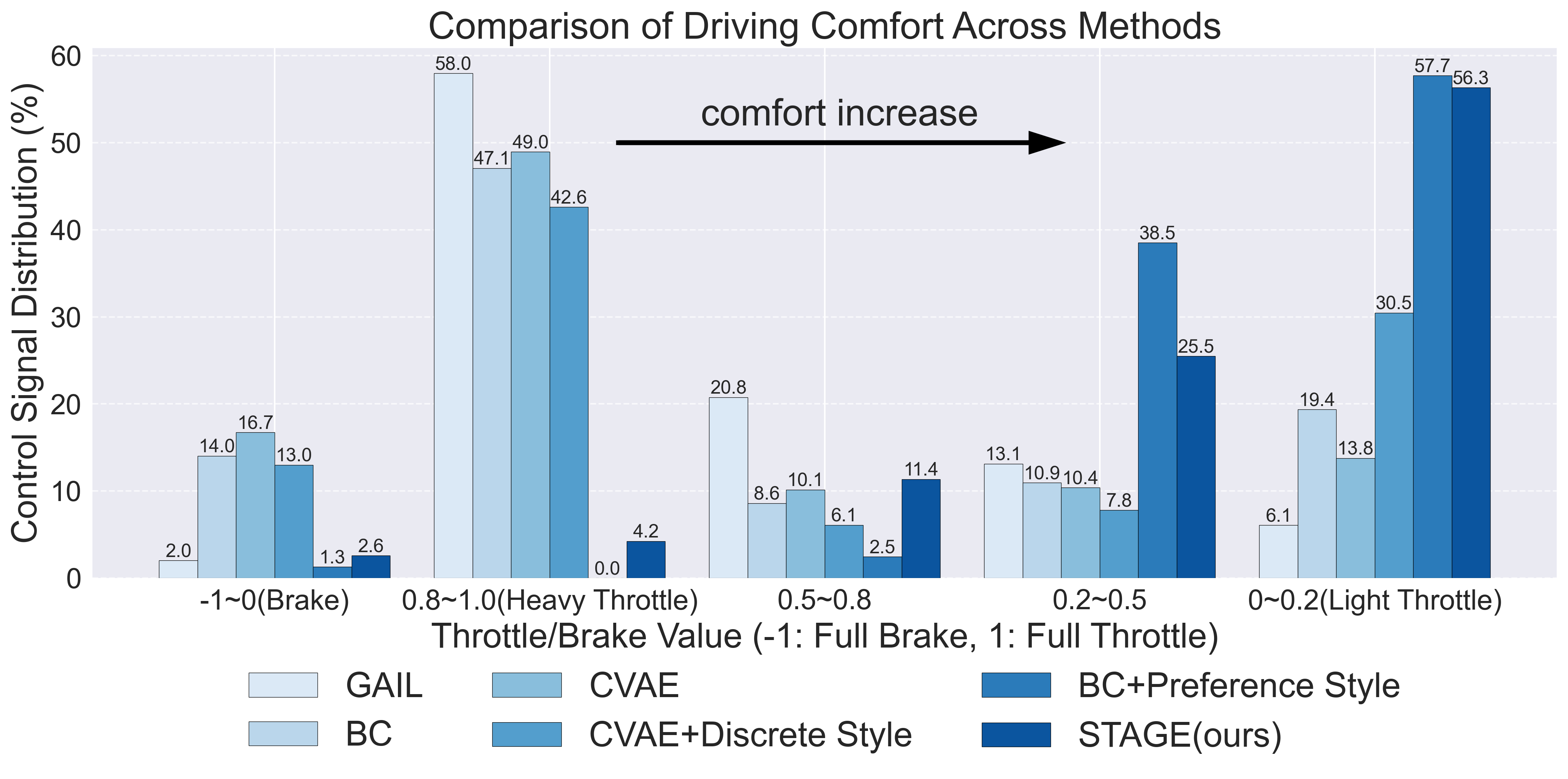}
      \caption{The figure illustrates a comparison of driving comfort derived from throttle and brake control signals. In the figure, brake values range from [-1, 0], while throttle values range from [0, 1]. Frequent braking and extreme acceleration can cause discomfort to the driver, thereby reducing trust in the safety of autonomous driving. A good driver typically maintains a steady speed with minimal throttle input.}
      \label{acceleration}
\end{figure}

\section{Conclusion}
We present STAGE, a human-machine collaborative driving system that generates trajectories based on driving style values. Built upon a multimodal imitation learning framework, STAGE employs a compressed encoding structure to capture action modalities, which are then decomposed into two components: driving style preferences and VAE latent variables. 
Driving style preferences are learned through aggressive scoring rules. The resulting style values go beyond the coarse-grained classification-based driving style recognition of previous methods, enabling fine-grained and monotonic correspondence to driving behaviors. 
And during execution, the human driver inputs a style value as a condition for human-like trajectory generation, enabling a low attention cost, human-intended self driving paradigm.

However, this work has several limitations. One notable challenge is aligning styles across different traffic scenarios. As observed in Figure \ref{scene}, driving trajectories generated with the same style value in different road scenarios may be perceived by the same driver as exhibiting varying levels of aggressiveness. 
For instance, the same style value could correspond to fully accelerating at the starting point or making a slow turn at an intersection. 
This discrepancy may require drivers to frequently adjust the style value across scenarios to meet their dynamic preferences. 
Addressing this issue may involve designing more comprehensive aggression scoring mechanisms that consider diverse driving scenarios.
Another limitation is how to efficiently learn multidimensional style values. As mentioned in Section III.A, while our method can theoretically be extended to more driving characteristics, learning a style value vector requires a larger data scale to accommodate combinations across dimensions, a condition we currently do not meet. Therefore, exploring more efficient style representation methods, such as natural language, would be beneficial. 
\textcolor{black}{This approach can also be better integrated with the widely used VLA methods, thereby enhancing the safety of model-driven driving.}
This challenge presents intriguing directions for future research.

\bibliographystyle{IEEEtranBST/IEEEtran}
\bibliography{IEEEtranBST/IEEEabrv,IEEEtranBST/IEEEexample}

@article{apollo_em,
  title={Baidu apollo em motion planner},
  author={Fan, Haoyang and Zhu, Fan and Liu, Changchun and Zhang, Liangliang and Zhuang, Li and Li, Dong and Zhu, Weicheng and Hu, Jiangtao and Li, Hongye and Kong, Qi},
  journal={arXiv preprint arXiv:1807.08048},
  year={2018}
}

@inproceedings{apollo_qp,
  title={Optimal vehicle path planning using quadratic optimization for baidu apollo open platform},
  author={Zhang, Yajia and Sun, Hongyi and Zhou, Jinyun and Pan, Jiacheng and Hu, Jiangtao and Miao, Jinghao},
  booktitle={2020 IEEE Intelligent Vehicles Symposium (IV)},
  pages={978--984},
  year={2020},
  organization={IEEE}
}

@INPROCEEDINGS{take_over,
  author={Kim, Jungsook and Kim, Hyun-Suk and Kim, Woojin and Yoon, Daesub},
  booktitle={2018 International Conference on Information and Communication Technology Convergence (ICTC)}, 
  title={Take-over performance analysis depending on the drivers’ non-driving secondary tasks in automated vehicles}, 
  year={2018},
  volume={},
  number={},
  pages={1364-1366},
  doi={10.1109/ICTC.2018.8539431}}

@article{drive_style_comfort,
  title={Comfort and Safety in Conditional Automated Driving in Dependence on Personal Driving Behavior},
  author={Vasile, Laurin and Dinkha, Naramsen and Seitz, Barbara and D{\"a}sch, Christoph and Schramm, Dieter},
  journal={IEEE Open Journal of Intelligent Transportation Systems},
  year={2023},
  publisher={IEEE}
}

@inproceedings{style_classification_raw,
  title={Driving style classification using long-term accelerometer information},
  author={Vaitkus, Vygandas and Lengvenis, Paulius and {\v{Z}}ylius, Gediminas},
  booktitle={2014 19th international conference on methods and models in automation and robotics (MMAR)},
  pages={641--644},
  year={2014},
  organization={IEEE}
}

@ARTICLE{style_classification_LMKNN,
  author={Tian, Xiang and Cai, Yingfeng and Sun, Xiaodong and Zhu, Zhen and Wang, Yong and Xu, Yiqiang},
  journal={IEEE Transactions on Transportation Electrification}, 
  title={Incorporating Driving Style Recognition Into MPC for Energy Management of Plug-In Hybrid Electric Buses}, 
  year={2023},
  volume={9},
  number={1},
  pages={169-181},
  doi={10.1109/TTE.2022.3181201}}

@article{style_implicit_gail,
  title={Modeling human driving behavior through generative adversarial imitation learning},
  author={Bhattacharyya, Raunak and Wulfe, Blake and Phillips, Derek J and Kuefler, Alex and Morton, Jeremy and Senanayake, Ransalu and Kochenderfer, Mykel J},
  journal={IEEE Transactions on Intelligent Transportation Systems},
  volume={24},
  number={3},
  pages={2874--2887},
  year={2022},
  publisher={IEEE}
}

@inproceedings{behavior_cloning1,
  title={A Framework for Behavioural Cloning.},
  author={Bain, Michael and Sammut, Claude},
  booktitle={Machine Intelligence 15},
  pages={103--129},
  year={1995}
}

@article{behavior_cloning2,
  title={Behavioral cloning from observation},
  author={Torabi, Faraz and Warnell, Garrett and Stone, Peter},
  journal={arXiv preprint arXiv:1805.01954},
  year={2018}
}

@article{behavior_cloning_drive,
  title={Chauffeurnet: Learning to drive by imitating the best and synthesizing the worst},
  author={Bansal, Mayank and Krizhevsky, Alex and Ogale, Abhijit},
  journal={arXiv preprint arXiv:1812.03079},
  year={2018}
}

@article{behavior_cloning_robot,
  title={Training robots without robots: Deep imitation learning for master-to-robot policy transfer},
  author={Kim, Heecheol and Ohmura, Yoshiyuki and Nagakubo, Akihiko and Kuniyoshi, Yasuo},
  journal={IEEE Robotics and Automation Letters},
  volume={8},
  number={5},
  pages={2906--2913},
  year={2023},
  publisher={IEEE}
}

@inproceedings{IRL,
  title={Algorithms for inverse reinforcement learning.},
  author={Ng, Andrew Y and Russell, Stuart and others},
  booktitle={Icml},
  volume={1},
  number={2},
  pages={2},
  year={2000}
}

@article{GAIL,
  title={Generative adversarial imitation learning},
  author={Ho, Jonathan and Ermon, Stefano},
  journal={Advances in neural information processing systems},
  volume={29},
  year={2016}
}

@article{DTrobot,
  title={Efficient spatiotemporal transformer for robotic reinforcement learning},
  author={Yang, Yiming and Xing, Dengpeng and Xu, Bo},
  journal={IEEE Robotics and Automation Letters},
  volume={7},
  number={3},
  pages={7982--7989},
  year={2022},
  publisher={IEEE}
}

@inproceedings{IBC,
  title={Implicit behavioral cloning},
  author={Florence, Pete and Lynch, Corey and Zeng, Andy and Ramirez, Oscar A and Wahid, Ayzaan and Downs, Laura and Wong, Adrian and Lee, Johnny and Mordatch, Igor and Tompson, Jonathan},
  booktitle={Conference on Robot Learning},
  pages={158--168},
  year={2022},
  organization={PMLR}
}

@article{aloha1,
  title={Learning fine-grained bimanual manipulation with low-cost hardware},
  author={Zhao, Tony Z and Kumar, Vikash and Levine, Sergey and Finn, Chelsea},
  journal={arXiv preprint arXiv:2304.13705},
  year={2023}
}

@article{diffusion_policy,
  title={The mechanisms of policy diffusion},
  author={Shipan, Charles R and Volden, Craig},
  journal={American journal of political science},
  volume={52},
  number={4},
  pages={840--857},
  year={2008},
  publisher={Wiley Online Library}
}

@inproceedings{uniad,
  title={Planning-oriented autonomous driving},
  author={Hu, Yihan and Yang, Jiazhi and Chen, Li and Li, Keyu and Sima, Chonghao and Zhu, Xizhou and Chai, Siqi and Du, Senyao and Lin, Tianwei and Wang, Wenhai and others},
  booktitle={Proceedings of the IEEE/CVF Conference on Computer Vision and Pattern Recognition},
  pages={17853--17862},
  year={2023}
}

@article{drivegpt4,
  title={Drivegpt4: Interpretable end-to-end autonomous driving via large language model},
  author={Xu, Zhenhua and Zhang, Yujia and Xie, Enze and Zhao, Zhen and Guo, Yong and Wong, Kwan-Yee K and Li, Zhenguo and Zhao, Hengshuang},
  journal={IEEE Robotics and Automation Letters},
  year={2024},
  publisher={IEEE}
}

@article{car_follow_style,
  title={Human-like car-following modeling based on online driving style recognition},
  author={Ma, Lijing and Qu, Shiru and Song, Lijun and Zhang, Junxi and Ren, Jie},
  journal={Electron. Res. Arch},
  volume={31},
  pages={3264--3290},
  year={2023}
}

@article{MHMM_style,
  title={Finite mixture of the hidden Markov model for driving style analysis},
  author={Ding, Lusa and Zhu, Ting and Wang, Yanli and Zou, Yajie},
  journal={Journal of advanced transportation},
  volume={2022},
  number={1},
  pages={4989947},
  year={2022},
  publisher={Wiley Online Library}
}

@article{DOP_style,
  title={A learning-based discretionary lane-change decision-making model with driving style awareness},
  author={Zhang, Yifan and Xu, Qian and Wang, Jianping and Wu, Kui and Zheng, Zuduo and Lu, Kejie},
  journal={IEEE transactions on intelligent transportation systems},
  volume={24},
  number={1},
  pages={68--78},
  year={2022},
  publisher={IEEE}
}

@article{DETR_moe_style_traj,
  title={Safety-balanced driving-style aware trajectory planning in intersection scenarios with uncertain environment},
  author={Wang, Xiao and Tang, Ke and Dai, Xingyuan and Xu, Jintao and Xi, Jinhao and Ai, Rui and Wang, Yuxiao and Gu, Weihao and Sun, Changyin},
  journal={IEEE Transactions on Intelligent Vehicles},
  volume={8},
  number={4},
  pages={2888--2898},
  year={2023},
  publisher={IEEE}
}

@article{SVM_style_hardware,
  title={An Embedded Driving Style Recognition Approach: Leveraging Knowledge in Learning},
  author={Zhang, Chaopeng and Wang, Wenshuo and Ju, Zhiyang and Chen, Zhaokun and Venture, Gentiane and Xi, Junqiang},
  journal={IEEE Transactions on Intelligent Vehicles},
  year={2024},
  publisher={IEEE}
}

@article{metadrive,
  title={Metadrive: Composing diverse driving scenarios for generalizable reinforcement learning},
  author={Li, Quanyi and Peng, Zhenghao and Feng, Lan and Zhang, Qihang and Xue, Zhenghai and Zhou, Bolei},
  journal={IEEE transactions on pattern analysis and machine intelligence},
  volume={45},
  number={3},
  pages={3461--3475},
  year={2022},
  publisher={IEEE}
}

@inproceedings{DETR_BC,
  title={Safe real-world autonomous driving by learning to predict and plan with a mixture of experts},
  author={Pini, Stefano and Perone, Christian S and Ahuja, Aayush and Ferreira, Ana Sofia Rufino and Niendorf, Moritz and Zagoruyko, Sergey},
  booktitle={2023 IEEE International Conference on Robotics and Automation (ICRA)},
  pages={10069--10075},
  year={2023},
  organization={IEEE}
}

@inproceedings{cvae_driving,
  title={Vehicle trajectory prediction using intention-based conditional variational autoencoder},
  author={Feng, Xidong and Cen, Zhepeng and Hu, Jianming and Zhang, Yi},
  booktitle={2019 IEEE Intelligent Transportation Systems Conference (ITSC)},
  pages={3514--3519},
  year={2019},
  organization={IEEE}
}

@article{cvae_classifier_style,
  title={Driving style-based conditional variational autoencoder for prediction of ego vehicle trajectory},
  author={Kim, Dongchan and Shon, Hyukju and Kweon, Nahyun and Choi, Seungwon and Yang, Chanuk and Huh, Kunsoo},
  journal={IEEE Access},
  volume={9},
  pages={169348--169356},
  year={2021},
  publisher={IEEE}
}

@article{VAE_compare1,
  title={Generating drawing/grinding trajectories based on hierarchical CVAE},
  author={Aita, Masahiro and Nogi, Yuya and Sugawara, Keito and Sakaino, Sho and Tsuji, Toshiaki},
  journal={Advanced Robotics},
  pages={1--11},
  year={2025},
  publisher={Taylor \& Francis}
}

@inproceedings{VAE_compare2,
  title={Newtonianvae: Proportional control and goal identification from pixels via physical latent spaces},
  author={Jaques, Miguel and Burke, Michael and Hospedales, Timothy M},
  booktitle={Proceedings of the IEEE/CVF conference on computer vision and pattern recognition},
  pages={4454--4463},
  year={2021}
}

\end{document}